\documentclass[10pt,twocolumn,letterpaper]{article}

\usepackage{iccv}
\usepackage{times}
\usepackage{epsfig}
\usepackage{graphicx}
\usepackage{amsmath}
\usepackage{amssymb}

\usepackage[breaklinks=true,bookmarks=false]{hyperref}

\iccvfinalcopy 

\def\iccvPaperID{****} 

\ificcvfinal\fi

\begin{document}

\title{An Alarm System for Segmentation Algorithm Based on Shape Model}

\author{\and
Fengze Liu$^1$, Yingda Xia$^1$, Dong Yang$^2$, Alan Yuille$^1$, Daguang Xu$^2$\\
$^1$Johns Hopkins University, $^2$NVIDIA Corporation\\
{}
}

\maketitle
\ificcvfinal\thispagestyle{empty}\fi

\begin{abstract}
   It is usually hard for a learning system to predict correctly on rare events that never occur in the training data, and there is no exception for segmentation algorithms. Meanwhile, manual inspection of each case to locate the failures becomes infeasible due to the trend of large data scale and limited human resource.
   Therefore, we build an alarm system that will set off alerts when the segmentation result is possibly unsatisfactory, assuming no corresponding ground truth mask is provided. One plausible solution is to project the segmentation results into a low dimensional feature space; then learn classifiers/regressors to predict their qualities. Motivated by this, in this paper, we learn a feature space using the shape information which is a strong prior shared among different datasets and robust to the appearance variation of input data.
   The shape feature is captured using a Variational Auto-Encoder (VAE) network that trained with only the ground truth masks. During testing, the segmentation results with bad shapes shall not fit the shape prior well, resulting in large loss values. Thus, the VAE is able to evaluate the quality of segmentation result on unseen data, without using ground truth. Finally, we learn a regressor in the one-dimensional feature space to predict the qualities of segmentation results. Our alarm system is evaluated on several recent state-of-art segmentation algorithms for 3D medical segmentation tasks. Compared with other standard quality assessment methods, our system consistently provides more reliable prediction on the qualities of segmentation results.
\end{abstract}

\begin{figure}[!t]
\begin{center}
    \includegraphics[width=7.7cm]{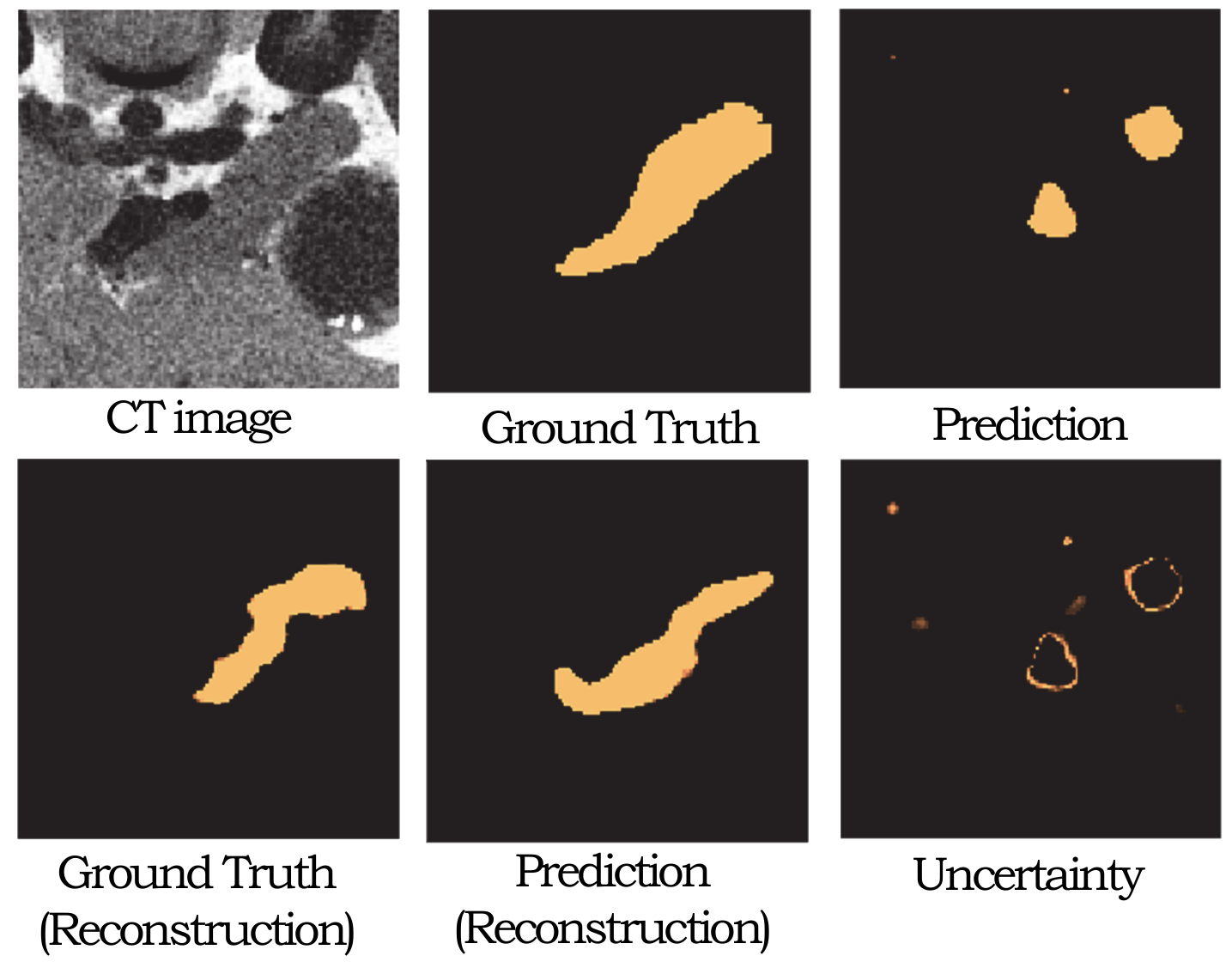}
\end{center}
\vspace{-5mm}
\caption{
   The visualize on an NIH CT data for pancreas segmentation. The Dice between GT and Prediction is $47.06$ (real Dice) while the Dice between Prediction and Prediction(Reconstruction) from VAE is $47.25$ (fake Dice). Our method uses the fake Dice to predict the former real Dice which is usually unknown at inference phase of real applications. This case shows how these two Dice scores are related to each other. In contrast, the uncertainty used in existing approaches (introduced in section $2$) mainly distributes on the boundary of predicted mask, which makes it a vague information when detecting the failure cases.
}
\label{Fig:motivation}
\vspace{-5mm}
\end{figure}
\section{Introduction}

Segmentation algorithms often fail on rare events, and it is hard to fully avoid such issue.  The rare events may occur due to limited number of training data. The most intuitive way to handle this problem is to increase the number of training data. However, the labelled data is usually hard to collect especially in medical domain, e.g., fully annotating a 3D medical CT scan requires professional radiology knowledge and several hours of work. Meanwhile, even large number of labelled data is usually unable to cover all possible cases. Previously, various methods have been proposed to make better use of the training data, like sampling strategies paying more attention to the rare events \cite{Yan18}. But still they may fail on rare events that never occur in the training data. Another direction is to increase the robustness of the segmentation algorithm to rare events. \cite{Unc17} proposed the Bayesian neural network that models the uncertainty as an additional loss to make the algorithm more robust to noisy data. These kinds of methods make the algorithm insensitive to certain types of perturbations, but the algorithms may still fail on other perturbations.

Since it is hard to completely prevent the segmentation algorithm from failure, we consider detecting the failure instead: build up an alarm system cooperating with the segmentation algorithm, which will set off alerts when the system finds that the segmentation result is not good enough. It is assumed that there is no corresponding ground truth mask, which is usually true after the model deployment due to the trend of large data scale and limited human resource. This task is also called as quality assessment. Several works have been proposed in this field. \cite{SNC18} applied Bayesian neural network to capture the uncertainty of the segmentation result and set off alarm based on its value. However, this system also suffers from rare events since the segmentation algorithms often make mistakes confidently on some rare events \cite{Xie17}, shown in Figure \ref{Fig:motivation}. \cite{EWOG12} provided an effective way by projecting the segmentation results into a feature space and learn from this low dimension space. They manually designed several heuristic features, e.g., size, intensity, and assumed such features would indicate the quality of the segmentation results.
After projecting the segmentation results into a low-dimensional feature space, they learned a classifier to predict its quality which distinguishes good segmentation results from bad ones directly.
In a reasonable feature space, the representation of the failure output should be far from that of the ground truth when the segmentation algorithm fails. So the main problems is what these ``good'' features are and how to capture them. Many features selected in \cite{EWOG12} are actually less related to the quality of segmentation results, e.g., size.

\begin{figure*}[!t]
\begin{center}
    \includegraphics[width=15cm]{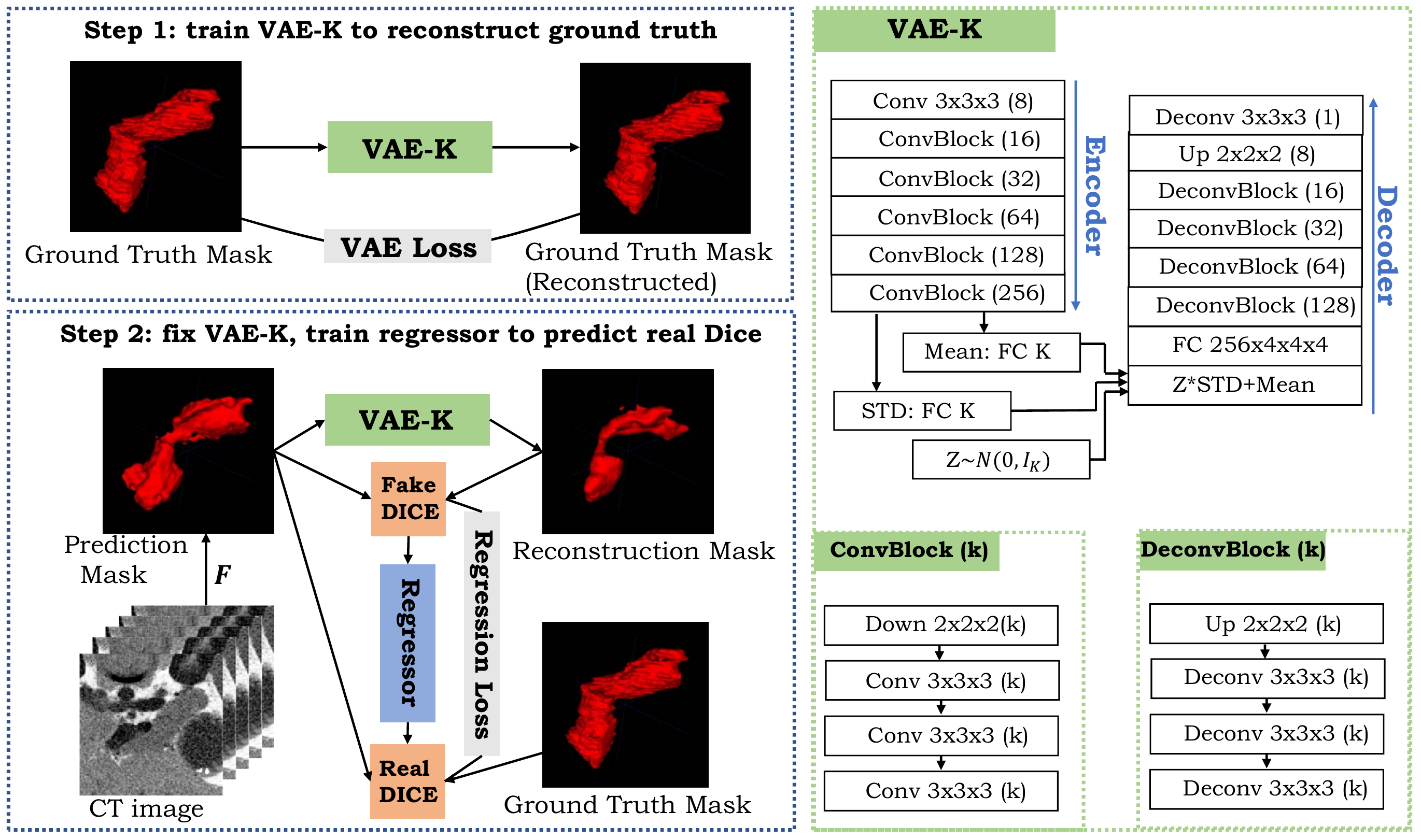}
\end{center}
\vspace{-0.3cm}
\caption{
   The architecture of our alarm system. In train step $1$, the VAE is trained to reconstruct the ground truth masks. In train step $2$, the parameters of VAE are fixed and a regressor is trained to predict the real Dice score. $F$ represents a preparation segmentation algorithm which is used to generate prediction masks for training the regressor. During testing, $F$ is replaced with the target segmentation algorithm to be evaluated. On the right side we show the structure of VAE used. (\textbf{Conv}: convolution layers with stride $1$. \textbf{Down}: convolution layers with stride $2$. \textbf{Deconv}: transpose convolution layers with stride $1$. \textbf{Up}: transpose convolution layers with stride $2$. \textbf{FC}: fully connected layers. $\pmb{k}$: convolution kernel numbers.) Further details about the structure are presented in section $4.3$.
}
\label{Fig:pipeline}
\vspace{-0.5cm}
\end{figure*}

In our system, we choose the shape feature which is more representative and robust because the segmented objects (foreground in the volumetric mask) usually have stable shapes among different cases even though their image appearance may vary a lot, especially in 3D. So the shape feature could provide a strong prior information for judging the quality of segmentation results, i.e., bad segmentation results tend to have bad shapes and vice versa. Furthermore, modeling the prior from the segmentation mask space is much easier than doing it in the image space. The shape prior can be shared among different datasets while the features like image intensity are affected by many factors. Thus, the shape feature can deal with not only rare events but also different data distributions in the image space, which shows great generalization power and potential in transfer learning. We propose to use the Variational Auto-Encoder(VAE) \cite{VAE14} to capture the shape feature. The VAE is trained on the ground truth masks, and afterwards we define the value of the loss function as the shape feature of a segmentation result when it is tested with VAE network. Intuitively speaking, after the VAE is trained, the bad segmentation results with bad shapes are just rare events to VAE because it is trained using only the ground truth masks, which are under the distribution of normal shapes. Thus they will have larger loss value. In this sense we are utilizing the fact that the learning algorithms will perform badly on the rare events. Formally speaking (detailed in Sec. $3.1$), the loss function, known as the variational lower bound, is optimized to approximate the function $\log{P(Y)}$ during the training process. So after the training, the value of the loss function given a segmentation result $\hat{Y}$ is close to $\log{P(\hat{Y})}$, thus being a good definition for the shape feature.

In this paper, we proposed a VAE-based alarm system for segmentation algorithms, shown in Figure \ref{Fig:pipeline}. The qualities of the segmentation results can be well predicted using
our system. To validate the effectiveness of our alarm system, we test it on multiple segmentation algorithms. These segmentation algorithms are trained on one dataset and tested on several other datasets to simulate when the rare events occur. The performance for the segmentation algorithms on the other datasets (rather than the training dataset) varies a lot but our system can still predict their qualities accurately. We compare our system with several other alarm systems on the above tasks and ours outperforms them by a large margin, which shows the importance of shape feature in the alarm system and the great power of VAE in capturing the shape feature.

\section{Related Work}
\paragraph{Quality Assessment:} \cite{Unc17} employed Bayesian neural network (BNN) to model the aleatoric and epistemic uncertainty. Afterwards, \cite{bnn18} applied the BNN to calculate the aleatoric and epistemic uncertainty on medical segmentation tasks. \cite{SNC18} utilized the BNN and model another kind of uncertainty-based on the entropy of segmentation results. They calculated a doubt score by summing over weighted pixel-vise uncertainty. 

Other methods like \cite{valindria2017reverse}\cite{RR} used registration based approach for quality assessment. It registered the image of testing case with a set of reference image and also transfer the registration to the segmentation mask to find the most matching one. However it can be slow to register with all the reference image especially in 3D. Also the registration based approach can hardly be transferred between datasets or modalities. \cite{chabrier2006unsupervised} and \cite{gao2017novel} used unsupervised methods to estimate the segmentation quality using geometrical and other features. However their application in medical settings is not clear. \cite{EWOG12} introduced a feature space of shape and appearance to characterize a segmentation. The shape features in their system contain volume size and surface area, which are not necessarily related with the quality of the segmentation results. Meanwhile, \cite{robinson2018real} tried a simple method using image-segmentation pairs to directly regress the quality. \cite{cnniqa} used the feature from deep network for quality assessment. 
\vspace{-0.3cm}
\paragraph{Anomaly Detection:} Quality assessment is also related with Out-of-Distribution (OOD) detection. Investigation related research papers can be found in \cite{noveltydetection}. Previous works in this field \cite{hendrycks2016baseline} \cite{liang2017enhancing} made use of the softmax output in the last layer of a classifier to calculate the out-of-distribution level. In our case, however, for a segmentation method, we can only get a voxel-wise out-of-distribution level using these methods. How to calculate the out-of-distribution level for the whole mask as an entity becomes another problem. In addition, the segmentation algorithm can usually predict most of background voxels correctly with a high confidence, making the out-of-distribution level on those voxels less representative. 
\vspace{-0.4cm}
\paragraph{Auto-Encoder:} Auto-Encoder(AE), as a way of learning representation of data automatically, has been widely used in many areas such as anomaly detection \cite{dae18}, dimension reduction, etc. Unlike \cite{Shapenet15} which needs to pre-train with RBM, AE can be trained following an end-to-end fashion. \cite{PN16} learned the shape representation from point cloud form, while we choose the volumetric form as a more natural way to corporate with segmentation task. \cite{oktay2018anatomically} utilizes AE to evaluate the difference between prediction and ground truth but not in an unsupervised way. \cite{jordan} explored shape features using AE. \cite{aeiqa} utilized the reconstruction error of brain MRI image by AE and \cite{gan} used GAN for anomaly detection but it is sometimes hard to generate a realistic image \eg abdominal CT scan. \cite{Retinal} used AE and a one-class SVM to identify anomalous regions in OCT images through unsupervised learning on healthy examples. Variational autoencoder(VAE) \cite{VAE14}, compared with AE, adds more constraint on the latent space, which prevents from learning a trivial solution \eg identity mapping. \cite{vaeiqa} applied VAE for anomaly detection on MNIST and KDD datasets. In this paper we employ VAE to learn the shape representation for the volumetric mask and use that for quality assessment task. 

\section{Our VAE-based Alarm System}
We first define our task formally. Denote the datasets as $(\mathcal{X},\mathcal{Y})$, where $\mathcal{Y}$ is the label set of $\mathcal{X}$. We divide $(\mathcal{X},\mathcal{Y})$ into training set $(\mathcal{X}_t,\mathcal{Y}_t)$ and validation set $(\mathcal{X}_v,\mathcal{Y}_v)$. Suppose we have a segmentation algorithm $F$ trained on $\mathcal{X}_t$. Usually we validate the performance of $F$ on $\mathcal{X}_v$ using $\mathcal{Y}_v$. Now we want to do this task without $\mathcal{Y}_v$. Formally, we try to find a function $L$ such that
\begin{equation}
    \mathcal{L}(F(X),Y)=L(F,X;\omega)
\end{equation}
where $\mathcal{L}$ is a function used to calculate the similarity of the segmentation result $F(X)$ respect to the ground truth $Y$, i.e., the quality of $F(X)$. How to design $L$ to take valuable information from $F$ and $X$, is the main question. Recall that the failure may happen when $X$ is a rare event. But to detect whether an image $X$ is within the distribution of training data is very hard because of the complex structure of image space. In uncertainty-based method \cite{SNC18} and \cite{bnn18}, the properties of $F$ are encoded by sampling its parameters and calculating the uncertainty of output. The uncertainty does help predict the quality but the performance strongly relies on $F$. It requires $F$ to have Bayesian structure, which is not in our assumption. Also for a well-trained $F$, the uncertainty will mainly distribute on the boundary of segmentation prediction. So we change the formulation above to 
\begin{equation}
  \mathcal{L}(F(X),Y)= L(F(X);\omega)
\end{equation}
By adding this constraint, we still take the information from $F$ and $X$, but not in a direct way. The most intuitive idea to do is directly training a regressor on the segmentation results to predict the quality. But the main problem is that the regression parameters trained with a certain segmentation algorithm $F$ highly relate with the distribution of $F(X)$, which varies from different $F$. 

Following the idea of \cite{EWOG12}, we develop a two-step method. Firstly we encode the segmentation result $F(X)$ into the feature space, denoting as $S(F(X);\theta)$. Secondly we learn from the feature space to predict the quality of $F(X)$. Finally it changes to 
\begin{equation}
  \mathcal{L}(F(X),Y)=L(S(F(X);\theta);\omega)
\end{equation}

\subsection{Shape Feature from Variational Autoencoder}
In the first step we learn a feature space of shape from Variational Autoencoder (VAE) trained with the ground masks $Y\in\mathcal{Y}_t$, \ie using $S(Y;\theta)$ to indicate how perfect the shape of $Y$ is. Here we define the shape of the segmentation masks as the distribution of the masks in volumetric form. We assume the normal label $Y$ obeys a certain distribution $P(Y)$. For a predictive mask $\hat{y}$, its quality should be related with $P(Y=\hat{y})$. Our goal is to estimate the function $P(Y)$ using $S(Y;\theta)$. Recall the theory of VAE, we hope to find an estimation function $Q(z)$ minimizing the difference between $Q(z)$ and $P(z|Y)$, where $z$ is the variable of the latent space we want encoding $Y$ into, i.e. optimizing
\begin{equation}
    \mathcal{KL}[Q(z)||P(z|Y)]=E_{z\sim Q}[\log{Q(z)-\log{P(z|Y)}}]
\end{equation}

$\mathcal{KL}$ is Kullback-Leibler divergence. By replacing $Q(z)$ with $Q(z|Y)$, finally it would be deduced to the core equation of VAE \cite{VAEt}.
\begin{align}
    &\log{P(Y)}-\mathcal{KL}[Q(z|Y)||P(z|Y)]\nonumber\\
    &=E_{z\sim Q}[\log{P(Y|z)}]-\mathcal{KL}[Q(z|Y)||P(z)]
\end{align}

where $P(z)$ is the prior distribution we choose for $z$, usually Gaussian, and $Q(z|Y),P(Y|z)$ correspond to encoder and decoder respectively. Once $Y$ is given, $\log{P(Y)}$ is a constant. So by optimizing the RHS known as variational lower bound of $\log{P(Y)}$, we optimize for $\mathcal{KL}[Q(z|Y)||P(z|Y)]$. Here however we are interested in $P(Y)$. By exchanging the second term in LHS with all terms in RHS in equation $(5)$, we rewrite the training process as minimizing
\begin{align}
    &  E_{Y\sim \mathcal{Y}_t}\ \mathcal{KL}[Q(z|Y)||P(z|Y)] \nonumber\\
    &= E_{Y\sim \mathcal{Y}_t}|\log{P(Y)}-S(Y;\theta)|
\end{align}

We choose $E_{z\sim Q}[\log{P(Y|z)}]-\mathcal{KL}[Q(z|Y)||P(z)]$ to be $S(Y;\theta)$. $S(Y;\theta)$ is the loss function we use for training VAE and the training process is actually learning the parameters $\theta$ to best fit $\log{P(Y)}$ over the distribution of $Y$. So after training VAE, $S(Y;\hat{\theta})$ becomes a natural approximation for $\log{P(Y)}$ where $\hat{\theta}$ is the learned parameter. So we can just use $S(Y;\hat{\theta})$ as our shape feature. In this method we use Dice Loss \cite{VNet16} when training VAE, which is widely used in medical segmentation task. The final form of $S$ is
\begin{align}
    S(Y;\theta)&=E_{z\sim \mathcal{N}(\mu(Y),\Sigma(Y))}\frac{2|g(z)\cdot Y|}{|Y|^2+|g(z)|^2}\nonumber\\
    &-\lambda\ \mathcal{KL}[\mathcal{N}(\mu(Y),\Sigma(Y))||\mathcal{N}(0,1)]
\end{align}
where encoder $\mu,\Sigma$ and decoder $g$ are controlled by $\theta$, and $\lambda$ is a coefficient to balance the two terms. The first term is the Dice's coefficient between $Y$ and $g(z)$, ranging from $0$ to $1$ and equal to $1$ if $Y$ and $g(z)$ are equal. 

\subsection{Shape Feature for Predicting Quality}
In the second step we regress on the shape feature to predict the quality. We assume that the shape feature is good enough to obtain reliable quality assessment because intuitively thinking, for a segmentation result $F(X)$, the higher $\log{P(F(X))}$ is, the better shape $F(X)$ is in, thus the higher  $\mathcal{L}(F(X),Y)$ is and vice versa. Formally, taking the shape feature in section $3.1$, we can predict the quality by learning $\omega$ such that
\begin{equation}
    \mathcal{L}(F(X),Y)=L(S(F(X);\hat{\theta});\omega)
\end{equation}
Here the parameter $\hat{\theta}$ is learned by training the VAE, using labels in the training data $\mathcal{Y}_t$, and is then fixed during train step two. We choose $L$ to be a simple linear model, so the energy function we want to optimize is
\begin{equation}
    E(S(F(X);\hat{\theta});a,b) = ||aS(F(X);\hat{\theta})+b-\mathcal{L}(F(X),Y)||^2
\end{equation}
We only use linear regression model because the experiments show strong linear correlation between the shape features and the qualities of segmentation results. $\mathcal{L}$ is the Dice's coefficient, i.e. $\mathcal{L}(F(X),Y)=\frac{2|F(X)\cdot Y|^2}{|F(X)|^2+|Y|^2}$.
\begin{figure}[!t]
\begin{center}
    \includegraphics[width=8.5cm]{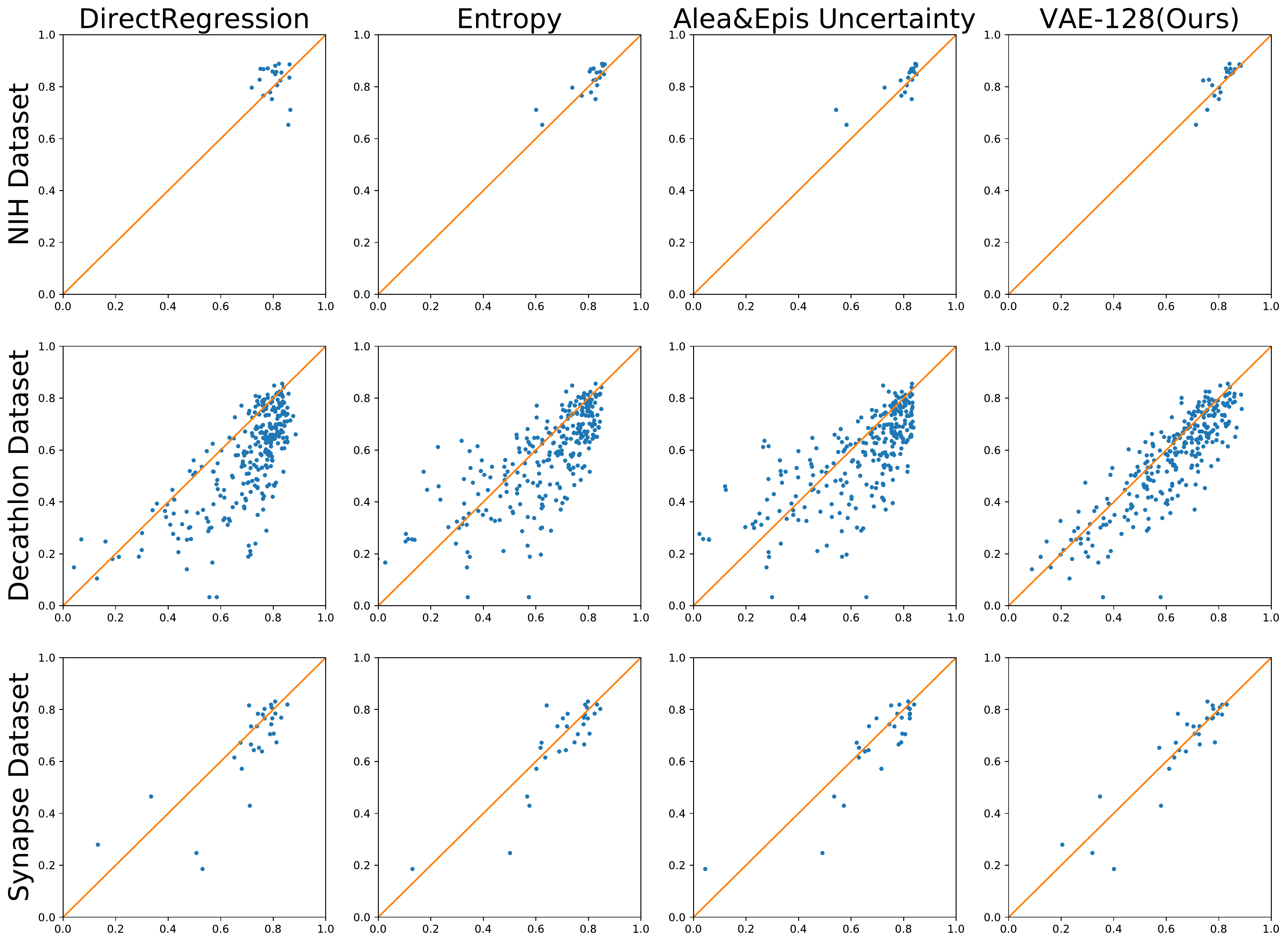}
\end{center}
\caption{
    This figure shows our predictive Dice score (x axis) vs real Dice score (y axis). For each row, the segmentation algorithm is tested on the left most dataset. The four figures in each row show how the segmentation results are evaluated by $4$ different methods.
}
\label{Fig:foo}
\end{figure}
\subsection{Training Strategy}
In step one, the VAE is trained only using labels in training data. Then in step two $\theta$ is fixed as $\hat{\theta}$. To learn $a,b$, the standard way is to optimize the energy function in $3.2$ using the segmentation results on the training data, i.e.
\begin{equation}
    \arg\min_{a,b} \sum_{(X,Y)\in(\mathcal{X}_t,\mathcal{Y}_t)}||aS(F(X);\hat{\theta})+b-\mathcal{L}(F(X),Y)||^2.
\end{equation}
Here the segmentation algorithm $F$ we use to learn $a,b$ is called the preparation algorithm. If $F$ is trained on $\mathcal{X}_t$, the quality of $F(X)$ would be always high, thus providing less information to regress $a,b$. To overcome this, we use jackknifing training strategy for $F$ on $\mathcal{X}_t$. We first divide $\mathcal{X}_t$ into $\mathcal{X}_t^1$ and $\mathcal{X}_t^2$. Then we train two versions of $F$ on $\mathcal{X}_t\setminus \mathcal{X}_t^1$ and $\mathcal{X}_t\setminus \mathcal{X}_t^2$ respectively, say $F_1$ and $F_2$. The optimizing function is then changed to 
\begin{align}
    &\arg\min_{a,b} \sum_{k=1,2}\sum_{(X,Y)\in(\mathcal{X}_t^k,\mathcal{Y}_t^k)}\nonumber\\
    &||aS(F_k(X);\hat{\theta})+b-\mathcal{L}(F_k(X),Y)||^2.
\end{align}

In this way we solve the problem above by simulating the performance of $F$ on the testing set. The most accurate way is to do leave-one-out training for $F$, but the time consumption is not acceptable, and two-fold split is effective enough according to experiments. When the training is done, we can test on any segmentation algorithm $G$ and data $X$ to predict the quality $Q = \hat{a}S(G(X);\hat{\theta})+\hat{b}$ where $\hat{a}$ and $\hat{b}$ are the learned parameters for step $2$ using the above strategy.

\section{Experimental Results}
In this section we test our alarm system on several recent algorithms for automatic pancreas segmentation that are trained on a public medical dataset. Our system achieves reliable predictions on the qualities of segmentation results. Furthermore, the alarm system remains effective when the segmentation algorithms are tested on other unseen datasets. We show better quality assessment capability and transferability compared with uncertainty-based methods and direct regression method. The quality assessment results are evaluated using mean absolute error (MAE), standard deviation of residual error (STD), Pearson correlation (P.C.) and  Spearman's correlation (S.C.) between the real quality (Dice's coefficient) and predictive quality. 

\begin{table*}[htbp]
    \centering
    \setlength{\tabcolsep}{1.7mm}
    \begin{tabular}{l c c c c c c c c c c c c}
\hline
\multicolumn{1}{l|}{}
& \multicolumn{4}{|c}{NIH Dataset}& \multicolumn{4}{|c}{MSD Dataset}&\multicolumn{4}{|c}{Synapse Dataset} \\
\multicolumn{1}{l|}{}  & MAE & STD & P.C.& \multicolumn{1}{c|}{S.C.}  & MAE & STD & P.C.& \multicolumn{1}{c|}{S.C.}  & MAE & STD & P.C.& \multicolumn{1}{c}{S.C.}\\
\hline\hline
\multicolumn{1}{l|}{Direct Regression} & 6.30&	7.93&	-18.36&	\multicolumn{1}{c|}{-1.50} & 14.47&	12.50&	72.26&	\multicolumn{1}{c|}{70.17}& 8.22&	10.82&	78.29&	\multicolumn{1}{c}{71.39}\\
\multicolumn{1}{l|}{Direct Regression+Image} & 11.74&	13.67&	2.13&	\multicolumn{1}{c|}{3.16}& 21.87&	20.83&	5.53&	\multicolumn{1}{c|}{9.22}& 13.80&	17.65&	36.83&	\multicolumn{1}{c}{39.80}\\
\multicolumn{1}{l|}{Jungo \etal \cite{SNC18}} & 3.51&	3.98&	82.21&	\multicolumn{1}{c|}{61.95}& 11.86&	16.31&	71.24&	\multicolumn{1}{c|}{77.71}& 9.45&	20.61&	73.32&	\multicolumn{1}{c}{79.93}\\
\multicolumn{1}{l|}{Kwon \etal \cite{bnn18}} & 4.07&	4.71&	\textbf{82.41}&	\multicolumn{1}{c|}{75.93}& 12.68&	18.31&	70.42&	\multicolumn{1}{c|}{77.77}& 9.77&	22.30&	74.80&	\multicolumn{1}{c}{81.13}\\
\hline\hline
\multicolumn{1}{l|}{VAE-2 \ \ \ \ \ \ \ \ (53.93)} & 5.31&	6.45&	56.66&	\multicolumn{1}{c|}{57.14}& 14.86&	10.73&	81.21&	\multicolumn{1}{c|}{77.63}& 9.63&	11.23&	79.66&	\multicolumn{1}{c}{68.19}\\
\multicolumn{1}{l|}{VAE-16 \ \ \ \ \ \ (72.46)} & 4.39&	4.84&	62.10&	\multicolumn{1}{c|}{76.69}& 9.83&	9.56&	84.86&	\multicolumn{1}{c|}{83.93}& 6.29&	8.30&	89.57&	\multicolumn{1}{c}{82.56}\\
\multicolumn{1}{l|}{VAE-128 \ \ \ \ (76.00)} & \textbf{2.89}&	\textbf{3.60}&	81.08&	\multicolumn{1}{c|}{\textbf{82.86}}& \textbf{8.14}&	\textbf{9.14}&	\textbf{86.23}&	\multicolumn{1}{c|}{85.02}& \textbf{4.93}&	\textbf{7.20}&	\textbf{90.92}&	\multicolumn{1}{c}{\textbf{86.07}}\\
\multicolumn{1}{l|}{VAE-1024 \  \ (79.65)} & 3.50&	4.15&	73.78&	\multicolumn{1}{c|}{80.90}& 8.42&	9.24&	85.81&	\multicolumn{1}{c|}{\textbf{85.17}}& 5.71&	8.00&	88.61&	\multicolumn{1}{c}{85.98}\\

\hline
\end{tabular}
    \setlength{\belowcaptionskip}{-20pt}

    \caption{Comparison between our method and baseline methods. The target segmentation (\ie BNN) algorithm is evaluated automatically without using ground truth. We have tried different structures for VAE (\eg VAE-128 for $128$-dimensional latent space). Of all the methods, VAE-128 achieves the highest performance. The numbers in brackets following the VAE methods are the average Dice score of reconstructing the ground truth masks on validation data. Usually with more accurate reconstruction of ground truth masks, the evaluation result is better but too accurate reconstruction may harm the evaluation capability (thinking of the identity mapping).}
    \label{tab:compare}
\vspace{-0.3cm}
\end{table*}

\subsection{Dataset and Segmentation Algorithm}
We adopt three public medical datasets and four recently published segmentation algorithms in total. All datasets consist of 3D abdominal CT images in portal venous phase with pancreas region fully annotated. The CT scans have resolutions of $512\times512\times h$ voxels with varying voxel sizes.
\begin{itemize}
\item \textbf{NIH Pancreas-CT Dataset (NIH)} The NIH Clinical Center performed 82 abdominal 3D CT scans\cite{NIH} from 53 male and 27 female subjects. The subjects are selected by radiologists from patients without major abdominal pathologies or pancreatic cancer lesions. 
\item \textbf{Medical Segmentation Decathlon (MSD)}\footnote{http://medicaldecathlon.com/index.html} The medical decathlon challenge collects 420 (281 Training +139 Testing) abdominal 3D CT scans from Memorial Sloan Kettering Cancer Center. Many subjects have cancer lesions within pancreas region.
\item \textbf{Synapse Dataset}\footnote{https://www.synapse.org/\#!Synapse:syn3193805/wiki/217789} The multi-atlas labeling challenge provides 50 (30 Training +20 Testing) abdomen CT scans randomly selected from a combination of an ongoing colorectal cancer chemotherapy trial and a retrospective ventral hernia study.
\end{itemize}

The testing data of the last two datasets is not used in our experiment since we do not have their annotations. The segmentation algorithms we choose are V-Net \cite{VNet16}, 3D Coarse2Fine \cite{3dc2f}, DeepLabv3 \cite{deeplab}, and 3D Coarse2Fine with Bayesian structure \cite{bnn18}. The first two algorithms are based on 3D networks while the DeepLab is 2D-based. The 3D Coarse2Fine with Bayesian structure is employed to compare with the uncertainty-based method, and we denote it as Bayesian neural network (BNN) afterwards.

\subsection{Baseline}
Our method is compared with three baseline methods. Two of them are based on uncertainty and the last one directly applies regression network on the prediction mask to regress quality in equation $(2)$:
\vspace{-0.1cm}
\begin{itemize}
    \item \textbf{Entropy Uncertainty}. \cite{SNC18} calculated the pixel-vise predictive entropy using Bayesian inference. Then, the uncertainty is summed up over the whole image to get the doubt score which would replace the shape feature in $(8)$ to regress the quality. The sum is weighted by the distance to predicted boundary, which somehow alleviates the bias distribution of uncertainty. Their method is done in 2D image and here we just transfer it to 3D image without essential difficulty.
    \item \textbf{Aleatoric and Epistemic Uncertainty}. \cite{bnn18} divided the uncertainty into two terms called aleatoric uncertainty and epistemic uncertainty. We implement both terms and calculate the doubt score in the same way as \cite{SNC18} because the original paper does not provide a way. The two doubt scores are used in predicting the quality.
    \item \textbf{Direct Regression}. A regression neural network is employed to directly learn the quality of predictive mask. It takes a segmentation mask as input and output a scalar for the predictive quality. 
\end{itemize}
\begin{table*}[]
    \footnotesize
    \centering
    \setlength{\tabcolsep}{3mm}
    \begin{tabular}{l c c c c c c c c c c}
\hline

\multicolumn{1}{c}{}
& \multicolumn{5}{|c}{3D Coarse2Fine}& 
\multicolumn{5}{|c}{3D VNet} \\
\multicolumn{1}{l|}{}  & MAE & STD & P.C.& S.C.& \multicolumn{1}{c|}{Dice}& MAE & STD & P.C.& S.C.&\multicolumn{1}{c}{Dice}\\
\hline\hline
\multicolumn{1}{l|}{NIH} & 3.46&	4.09&	89.95& 85.41 &	\multicolumn{1}{c|}{79.38} & 2.57&	3.24&	91.35&	84.51 & \multicolumn{1}{c}{81.21} \\
\multicolumn{1}{l|}{MSD} & 10.02&	9.45&	89.67&87.54&	\multicolumn{1}{c|}{51.88}& 9.34&	9.60&	86.52& 82.50&	\multicolumn{1}{c}{55.90} \\
\multicolumn{1}{l|}{Synapse} & 6.24&	9.00&	92.39&	84.29&\multicolumn{1}{c|}{62.10}& 5.67&	7.28&	91.65&80.11&	\multicolumn{1}{c}{64.93} \\
\hline\hline
\multicolumn{1}{c}{}
& \multicolumn{5}{|c}{DeepLabV3}& 
\multicolumn{5}{|c}{BNN} \\
\multicolumn{1}{l|}{}  & MAE & STD & P.C.& S.C.&\multicolumn{1}{c|}{Dice}& MAE & STD & P.C.& S.C.&\multicolumn{1}{c}{Dice}\\
\hline\hline
\multicolumn{1}{l|}{NIH} & 5.35&	5.83&	63.34&	78.80&\multicolumn{1}{c|}{81.53} & 2.89&	3.60&	81.08&	82.86&\multicolumn{1}{c}{82.15}\\
\multicolumn{1}{l|}{MSD} & 9.34&	9.60&	86.52&	82.50&\multicolumn{1}{c|}{54.96}& 8.14&	9.14&	86.23&85.02&	\multicolumn{1}{c}{57.10}\\
\multicolumn{1}{l|}{Synapse} & 5.67&	7.28&	91.65&80.11&	\multicolumn{1}{c|}{61.03} & 4.93&	7.20&	90.92&	86.07&\multicolumn{1}{c}{66.36}\\
\hline
\end{tabular}
    \setlength{\belowcaptionskip}{-20pt}
    \caption{Results of different target segmentation algorithms are evaluated by our alarm system on different datasets. The Dice column means the average Dice score for the segmentation algorithm tested with groundtruth on different datasets, provided for reference. Our system achieves comparable performance as in Table \ref{tab:compare} (see also in the right bottom cell) although the segmentation performance differs a lot between datasets.  Without tuning parameters, our alarm system can be directly applied to evaluate other segmentation algorithms}
    \label{tab:transfer}
    \vspace{-0.2cm}
\end{table*}
\vspace{-0.3cm}
\subsection{Implementation Detail}
The structure of VAE is shown in Figure \ref{Fig:pipeline}. We apply instance normalization on each convolution layer. The ReLU activation is applied on each layer except for the fully connected layer for mean value and the output layer is activated using the sigmoid function. The structure we use in the direct regression method is the encoder part of the VAE so that they are fair for comparison.

For data pre-processing, since the voxel size varies from case to case, which would affect the shape of pancreas and prediction of segmentation, we first re-sample the voxel size of all CT scans and annotation mask to 1$mm\times$1$mm\times$1$mm$. For training VAE, we apply simple alignment on the annotation mask. We employ a cube bounding box which is large enough to contain the whole pancreas region, centered at the pancreas centroid, then crop both volume and label mask out and resize it to a fixed size $128\times128\times128$. We only employ a simple alignment because the human pose is usually fixed when taking CT scans, e.g. stance, so that the organ will not rotate or deform heavily. For a segmentation prediction, we also crop and resize the predictive foreground to $128\times128\times128$ and feed it into VAE to capture the shape feature. 
\vspace{-0.1cm}
\begin{figure*}[!t]
\begin{center}
    \includegraphics[width=14cm]{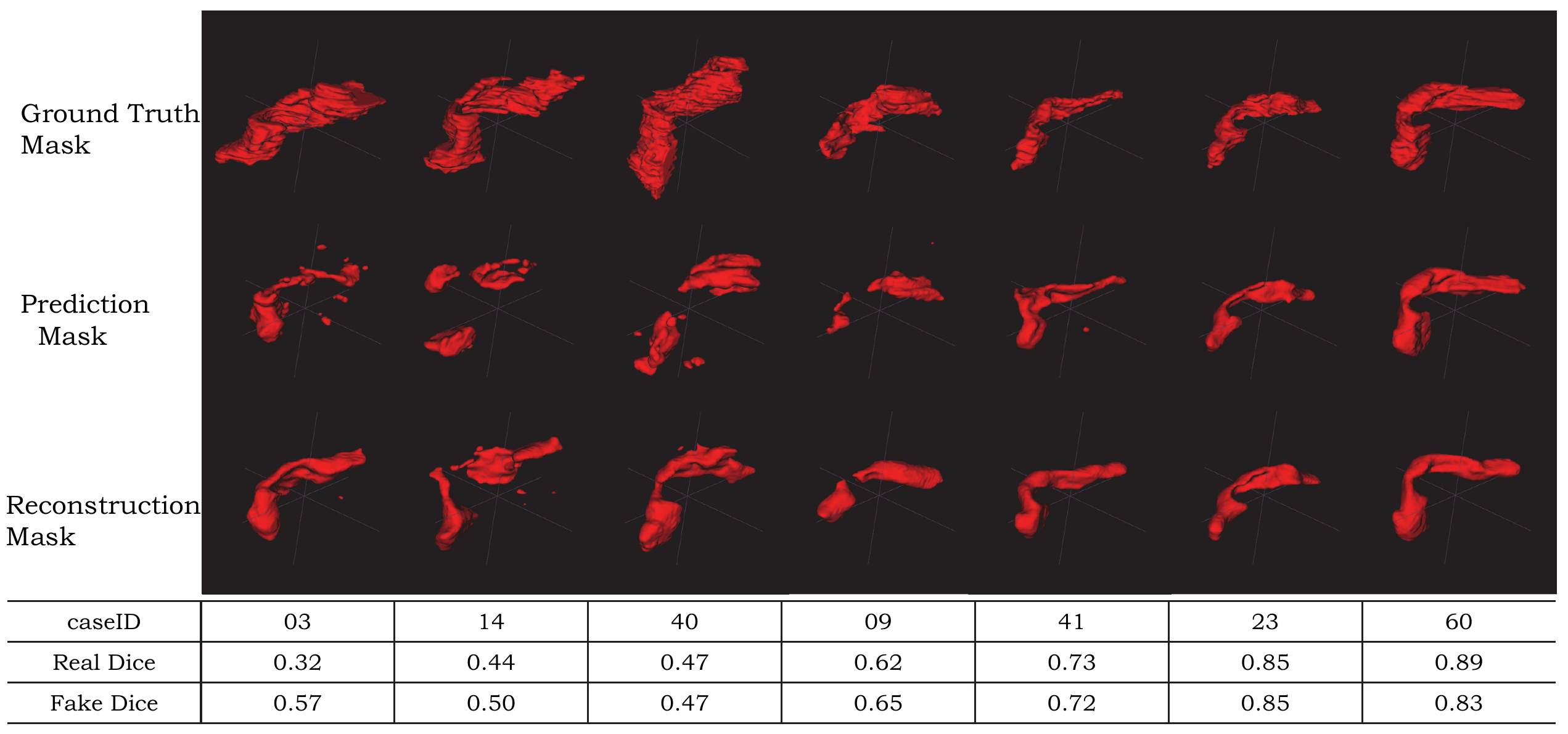}
\end{center}
\vspace{-3mm}
\caption{
   We visualize the performance of our evaluation system on different qualities of segmentation results. The real Dice score increases from left to right. The fake Dice score is highly correlated with the real Dice so that we can get good prediction of real Dice by applying simple regressor on the fake Dice.
}
\label{Fig:visual}
\vspace{-0.4cm}
\end{figure*}

During the training process, we employ rotation for $-10$, $0$, and $10$ degree along x,y,z axes($27$ conditions in total) and random translation for smaller than $5$ voxel on annotation mask as data augmentation. This kind of mild disturbance can enhance the data distribution but keep the alignment property of our annotation mask. We tried different dimension of latent space and finally set it to $128$. We found that VAE with latent space of different dimension will have different capability in quality assessment. The hyper parameter $\lambda$ in object function of VAE is set to $2^{-5}$ to balance the small value of Dice Loss and large KL Divergence. We trained our network by SGD optimizer. The learning rate for training VAE is fixed to $0.1$. Our framework and other baseline models are built using TensorFlow. All the experiments are run on NVIDIA Tesla V100 GPU. The first training step is done in total 20000 iterations and takes about $5$ hours.

\subsection{Primary Results and Discussion}
We split NIH data into four folds and three of them are used for training segmentation algorithms and VAE; the remaining one fold, together with all training data from MSD and Synapse datasets forms the validation data to evaluate our evaluation method. First we learn the parameter of VAE using the training label of NIH dataset. Then we choose BNN as the preparation algorithm mentioned in section $3.3$. The training strategy in section $3.3$ is applied on it to learn the parameters of regression. For all the baseline methods, we employ the same training strategy of jackknifing as in our method and choose the BNN as preparation algorithm for fair comparison. Finally we predict the quality of segmentation mask on the validation data for all the segmentation algorithms. Note that all segmentation algorithms are trained only on the NIH training set. 

Table \ref{tab:compare} compared our method and three baselines by assessing the BNN segmentation result of validation datasets. In general, our method achieves the lowest error and variance on all datasets. In our experiment, the preparation algorithm BNN achieves $82.15$, $57.10$ and $66.36$ average Dice score tested on NIH, MSD and Synapse datasets respectively. The segmentation algorithm trained on NIH will fail on some cases of other datasets, and our alarm system still works well without tuning the parameters of VAE and regressor on other datasets. More detailed result is as shown in Figure \ref{Fig:foo}. We can clearly observe that our method provides more accurate quality assessment result. For uncertainty-based methods, as shown in Figure \ref{Fig:motivation}, the uncertainty often distributes on the boundary of predicted masks but not on the missing parts or false positive parts and the transferability is not strong since it relies on the segmentation algorithm. For direct regression method, we use the encoder part of VAE-1024 followed by a 2-layer fully connection. The training data of direct regression method is the augmentated testing data of $F_1$, $F_2$ on $\mathcal{X}_t^1$, $\mathcal{X}_t^2$ respectively as in section $3.3$. So the number of training data for direct regression method is the same as ours but our method shows better capability of predicting the quality.

Table \ref{tab:transfer} shows the quality assessment results of our method for $4$ different segmentation algorithms. The result of BNN is better because the preparation algorithm we use for training the regressor is also BNN. Without tuning parameters, our method remains reliable when the segmentation algorithms to be evaluated and the dataset to be tested on are changed, which shows strong transferability. \begin{table*}[]
    \centering
    \begin{tabular}{l c c c c c c c c}
\hline
\multicolumn{1}{l|}{}
& \multicolumn{4}{c}{MSD Dataset Pancreas}& \multicolumn{4}{|c}{MSD Dataset Tumor} \\
\multicolumn{1}{l|}{}  & MAE & STD & P.C.& \multicolumn{1}{c|}{S.C.}  & MAE & STD & P.C.& \multicolumn{1}{c}{S.C.}  \\
\hline\hline
\multicolumn{1}{l|}{Direct Regression} & 7.48&	8.64&	56.48&	\multicolumn{1}{c|}{44.49} & 23.20&	29.81&	45.50&	\multicolumn{1}{c}{45.36}\\
\multicolumn{1}{l|}{Jungo \etal \cite{SNC18}} & 7.24&	8.79&	54.38&	\multicolumn{1}{c|}{49.29}& 26.57&	29.78&	-23.87&	\multicolumn{1}{c}{-20.23}\\
\multicolumn{1}{l|}{Kwon \etal \cite{bnn18}} & 6.94&	8.54&	62.15&	\multicolumn{1}{c|}{\textbf{61.20}}& 26.14&	29.24&	14.61&	\multicolumn{1}{c}{14.70}\\
\multicolumn{1}{l|}{VAE-1024(Ours)} &\textbf{6.03}&	\textbf{7.63}&	\textbf{68.40}&	\multicolumn{1}{c|}{59.65}& \textbf{20.21}&	\textbf{23.60}&	\textbf{60.24}&	\multicolumn{1}{c}{\textbf{63.30}}\\

\hline
\end{tabular}
    \setlength{\belowcaptionskip}{-10pt}
    \caption{Results for evaluating both pancreas and tumor segmentation. The MAE number for pancreas is better than those in Table \ref{tab:compare} since there are more training samples in the MSD dataset. For tumor evaluation, all the methods are not doing well but our method reveal the strongest correlation between the real quality and the predictive quality. Since detecting tumor itself is a very hard task, the segmentation prediction for tumor is often with more variance. The alarm system needs more careful design to deal with that big variance.}
    \label{tab:tumor}
\vspace{-0.4cm}
\end{table*}
\vspace{-0.4cm}
\paragraph{Why it works:} In the experiments we use $S(F(X);\hat{\theta})$ as the input of regressor. However we find the second term of $S(F(X);\hat{\theta})$ is less related with the real Dice (So in Figure \ref{Fig:pipeline} we only put the fake Dice there, which is the first term of $S(F(X);\hat{\theta})$). That means VAE can encode masks with bad shape into normal points in the latent space so that the reconstructions are of normal shape, which makes the fake Dice low. We visualize some cases in Figure \ref{Fig:visual} for showing this property of VAE. For bad segmentation predictions, the reconstruction masks from VAE indeed look more like a pancreas.

\subsection{Ablation Experiments}
We also run ablation experiments for different structures of VAE and for evaluating foreground without strong shape prior, \ie tumor region. 

\vspace{-0.4cm}
\paragraph{Different VAE Structures:} Table \ref{tab:compare} also shows results of VAE with latent space of different dimensions. With bigger latent space, VAE can reconstruct the ground truth masks better which generally indicates stronger evaluation capability. But for VAE-1024, the reconstruction is the best but the prediction result is not as good as VAE-128. We have also tried larger latent space like VAE-10000, and it can reconstruct the ground truth masks almost perfectly. But it is more like an identity mapping, making it impossible for the evaluation task.
\vspace{-0.5cm}
\paragraph{Combine With Texture:} Since our alarm system only uses the information of segmentation masks, the texture information, which can be important in evaluating the segmentation quality, is missing. We tested it with a very intuitive setting, \ie, for the direct regression method, we concatenate the image and segmentation masks together and use that as input for training the regression network. The result is shown in Table \ref{tab:compare}  ``Direct Regression+Image". We see that with the same number of training data, the performance is even worse than only taking the segmentation mask as input. We think it is because the complex structure of image will confuse the regression network for learning the quality. \cite{gan} and \cite{aeiqa} developed textured based methods on OCT and brain MRI data respectively, while in our experiments, it is hard to generate realistic abdominal CT scans. So how to better combine the texture with the segmentation mask is another direction worth exploring.
\vspace{-0.5cm}
\paragraph{Evaluate Object With Large Shape Variance:} We also compare baseline methods and our method on evaluating segmentation of object with less stable shape \eg tumor. The MSD dataset also provides voxel-wised label of pancreatic tumor. Instead of only evaluating the tumor prediction (requires accurate localization of tumor bounding boxes which is a hard task already), we evaluate both the tumor and pancreas segmentation at the same time so that we can use the bounding box of pancreas. Since this is a multi-class problem now, we adapt the VAE to take the one-hot encoding segmentation masks as input and change the original Dice loss to multi-class Dice loss. Similarly, we adapt the baseline methods so that they can fit in this multi-class evaluating problem. For direct regression method, it is trained to regress pancreas Dice score and tumor Dice score at the same time. For uncertainty-based method, uncertainty for both pancreas and tumor are calculated. We randomly split the MSD dataset into two parts and one is used for training while the other one for validation. For the training process we still apply the strategy as in section $3.3$. We also train a BNN for pancreas and tumor segmentation as the target algorithm to evaluate and it reaches $72.52$ and $35.34$ average Dice score on pancreas and tumor respectively. The detailed comparison is shown in Table \ref{tab:tumor}. For the uncertainty-based method, the tumor segmentation evaluation is quite bad because the segmentation algorithm often wrongly segments the tumor confidently, which also proves the limitation of uncertainty-based method on quality assessment. For the direct regression method, as there are more training data ($60\rightarrow 140$ before augmentation), the number is better than that in Table \ref{tab:compare}, which is common for a learning system. Our method still performs the best although it is not satisfactory, as there are many cases with $0$ Dice score on tumor segmentation which are hard to predict the quality only from the segmentation mask. Note that the correlation between the real quality and predictive quality of our method is much stronger, which means even with weak shape prior, our method can still capture some useful information from the segmentation mask.

\vspace{-0.1cm}
\section{Conclusion}
\vspace{-0.2cm}

In the paper we presented a VAE based alarm system for segmentation algorithms which predicts the qualities of the segmentation results without using ground truth. We claim that the shape feature is useful in predicting the qualities of the segmentation results. To capture the shape feature, we first train a VAE using ground truth masks. We utilize the fact that rare events usually achieve larger loss value, and successfully detect the out-of-distribution shape according to the loss value in the testing time. In the second step we collect the segmentation results of the segmentation algorithm on the training data, and extract the shape feature of them to learn the parameters of regression. By applying jackknifing training on the preparation algorithm we can obtain more accurate regression parameters.

Our proposed method outperforms the standard uncertainty-based methods and direct regression methods, and possesses better transferability to other datasets and other segmentation algorithms. The reliable quality assessment results prove both that the shape feature capturing from VAE is meaningful and that the shape feature is useful for quality assessment in the segmentation task.
{\small
\bibliographystyle{ieee_fullname}
\bibliography{egbib}
}

\end{document}